\documentclass[sigconf]{acmart}

\usepackage{tabularx}
\usepackage{multirow}
\usepackage{graphicx}
\usepackage{float}

\newcolumntype{Y}{>{\centering\arraybackslash}X}

\AtBeginDocument{%
  }

\copyrightyear{2025}
\acmYear{2025}
\setcopyright{acmlicensed}\acmConference[MAD'25]{4th ACM International
Workshop on Multimedia AI against Disinformation }{June 30-July 3
2025}{Chicago, IL, USA}
\acmBooktitle{4th ACM International Workshop on Multimedia AI against
Disinformation (MAD'25), June 30-July 3 2025, Chicago, IL, USA}
\acmDOI{10.1145/3733567.3735572}
\acmISBN{979-8-4007-1891-5/2025/06}

\begin{document}


\title{Do DeepFake Attribution Models Generalize?}

\author{Spiros Baxavanakis}
\email{sprbax@gmail.com}
\affiliation{%
  \institution{Information Technologies Institute @ CERTH}
      \city{Thessaloniki}
  \country{Greece}
}

\author{Manos Schinas}
\email{manosetro@iti.gr}
\orcid{0000-0000-0000}
\affiliation{%
  \institution{Information Technologies Institute @ CERTH}
    \city{Thessaloniki}
  \country{Greece}
}

\author{Symeon Papadopoulos}
\email{papadop@iti.gr}
\orcid{0000-0000-0000}
\affiliation{%
  \institution{Information Technologies Institute @ CERTH}
      \city{Thessaloniki}
  \country{Greece}
}


\begin{abstract}

Recent advancements in DeepFake generation, along with the proliferation of open-source tools, have significantly lowered the barrier for creating synthetic media. This trend poses a serious threat to the integrity and authenticity of online information, undermining public trust in institutions and media. State-of-the-art research on DeepFake detection has primarily focused on binary detection models. A key limitation of these models is that they treat all manipulation techniques as equivalent, despite the fact that different methods introduce distinct artifacts and visual cues. Only a limited number of studies explore DeepFake attribution models, although such models are crucial in practical settings. By providing the specific manipulation method employed, these models could enhance both the perceived trustworthiness and explainability for end users. In this work, we leverage five state-of-the-art backbone models and conduct extensive experiments across six DeepFake datasets. First, we compare binary and multi-class models in terms of cross-dataset generalization. Second, we examine the accuracy of attribution models in detecting \emph{seen} manipulation methods in unknown datasets, hence uncovering  data distribution shifts on the same DeepFake manipulations. Last, we assess the effectiveness of contrastive methods in improving cross-dataset generalization performance. Our findings indicate that while binary models demonstrate better generalization abilities, larger models, contrastive methods, and higher data quality can lead to performance improvements in attribution models. The code of this work is available on \href{https://github.com/mever-team/deepfake-contrastive-attribution}{GitHub}.
\end{abstract}






\maketitle

\section{Introduction}\label{sec:intro}

\begin{figure*}[!htbp]
    \centering
    \small
    \includegraphics[width=0.7\textwidth]{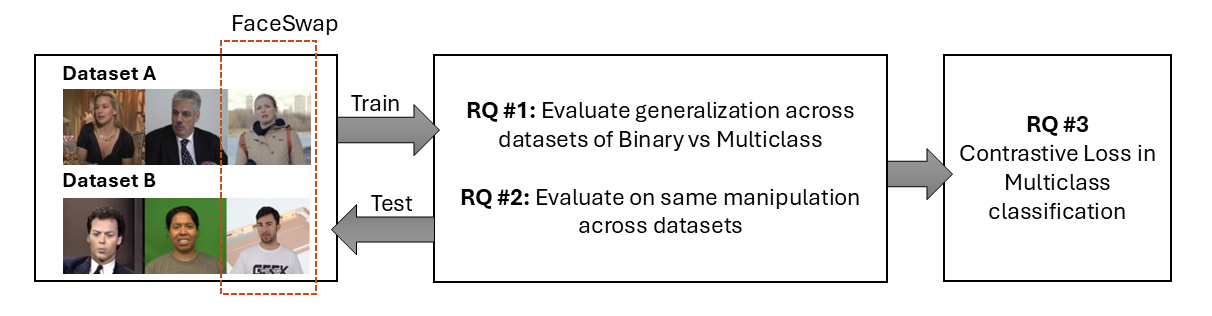}
    \caption{Overview of the research methodology addressing RQ1, RQ2, and RQ3.}
    \label{fig:methodology}
\end{figure*}

Recently, deep neural networks have significantly advanced image and video editing capabilities, with their use spreading across several prominent multimedia software products. 
While this technology enhances productivity and creativity, it has also been exploited for malicious purposes. In particular, deep learning-based approaches, known as DeepFakes, are used to modify a person's facial expressions \cite{nirkin2022fsganv2}, identity \cite{li2019faceshifter}, or facial attributes \cite{lin2021anycost}. Given the widespread availability of open source implementations \cite{DBLP:journals/pami/NirkinKH23,DBLP:journals/pr/LiuPGCZZ23,DBLP:conf/iccv/ShioharaYT23} and the numerous online DeepFake  services\footnote{For instance: \href{https://hoodem.com/}{Hoodem}, \href{https://deepfakesweb.com/}{DeepfakesWeb}, \href{https://www.facemagic.ai}{FaceMagic}, \href{https://reface.ai}{ReFace}, \href{https://www.vidnoz.com}{VidNoz}}, it is straightforward to create harmful realistic video content. 
As a result, DeepFake detection has gained significant research attention, with numerous methods proposed for images \cite{DBLP:conf/cvpr/LiBZYCWG20,DBLP:conf/cvpr/ZhaoZ0WZY21,DBLP:conf/cvpr/ShioharaY22}, videos \cite{DBLP:conf/mm/ZiCCMJ20,DBLP:conf/cvpr/WangBZWL23,DBLP:conf/cvpr/HaliassosVPP21}, and audio-visual content \cite{DBLP:conf/cvpr/FengCO23,DBLP:conf/mm/ChughGDS20,DBLP:conf/mm/MittalBCBM20}. Although current methods achieve strong within-dataset performance, they fall short in terms of generalization ability as shown by cross-dataset and out-of-distribution experiments \cite{DBLP:journals/corr/abs-2401-04364}. 

Beyond generalization, there is an increasing demand for trustworthiness in deep learning systems. One key aspect that can enhance the credibility of DeepFake detectors is manipulation identification or attribution. This involves producing not only a manipulation probability but also identifying the specific technique used to alter the image or video. However, manipulation attribution remains an underexplored area, with only a limited number of studies addressing this challenging problem \cite{DBLP:conf/mmsp/JainKM21,DBLP:conf/icip/JiaLL22,DBLP:journals/tcsv/LiNYFZ23,si2023efficient}, in contrast to the large body of work treating DeepFake detection as a binary classification task \cite{DBLP:journals/corr/abs-2401-04364}. Yet, formulating the problem as binary classification is limited, as it overlooks the substantial differences between manipulation techniques that often require distinct detection strategies.

To adopt manipulation attribution models in practice, it is crucial to compare their performance and reliability against binary models. A reliable attribution model should not only detect manipulations but also correctly identify known manipulation types across different datasets. Without this capability, such models are ill-suited for the diverse and dynamic nature of real-world DeepFake detection. This motivates the following three research questions:

\begin{enumerate}
    \item How do attribution and binary models compare in terms of cross-dataset generalization (\textbf{RQ1})?
    \item Do attribution models maintain same-manipulation performance in cross dataset settings (\textbf{RQ2})?
    \item Do contrastive methods in training improve attribution performance (\textbf{RQ3})?
\end{enumerate}

To answer \textbf{RQ1}, we train both binary and multi-class (attribution) classification models using established backbone architectures on several widely used DeepFake datasets. We then conduct cross-dataset evaluations to compare generalization performance under the two training paradigms. This lays the groundwork for addressing \textbf{RQ2} by facilitating cross-dataset evaluations \textit{exclusively} on manipulation methods shared between train and test datasets. Our aim in answering \textbf{RQ2} is to examine any inherent cross-dataset distribution shifts characteristic of DeepFakes generated using the same manipulation method. 
To address \textbf{RQ3}, we train attribution models using contrastive learning techniques, such as Triplet, NT-Xent, and Supervised Contrastive Loss, and compare their performance against standard attribution models. Answering \textbf{RQ3} is crucial for developing reliable DeepFake attribution models, as prior work suggests that contrastive learning can significantly improve performance in computer vision tasks \cite{DBLP:conf/nips/KhoslaTWSTIMLK20,DBLP:conf/icml/ChenK0H20}. Figure \ref{fig:methodology} depicts the overall approach. 

In summary, our contributions include the following:
\begin{itemize}
    \item We conduct a comprehensive comparison between binary and multi-class DeepFake detection models. Our findings demonstrate that binary models outperform multi-class models in terms of cross-dataset generalization, highlighting the need for improved generalization strategies in DeepFake attribution.
    \item We investigate the same-manipulation performance of DeepFake attribution models across different datasets. We identify significant accuracy drops when models are tested on seen manipulations from unknown datasets. This reveals the limitations of current attribution models in handling dataset distribution shifts.
    \item We explore the impact of contrastive learning methods on the performance of DeepFake attribution models. Our results indicate that while contrastive methods offer minimal performance gains for smaller networks, they significantly improve in terms of generalization on larger models.
\end{itemize}

\section{Related Work}
Since the rise of DeepFakes, researchers have largely focused on DeepFake detection \cite{DBLP:journals/ijcv/Juefei-XuWHGML22,DBLP:journals/corr/abs-2202-06228}. Some approaches leverage  physiological signals \cite{DBLP:journals/corr/abs-1901-02212,DBLP:conf/mm/QiGJXMFLZ20,DBLP:conf/wifs/LiCL18}, while others emphasize the importance of multimodal inputs to build robust DeepFake detectors \cite{DBLP:conf/mm/MittalBCBM20,DBLP:conf/mm/ChughGDS20}.
Recently, authors in \cite{DBLP:journals/corr/abs-2401-04364} evaluated 51 DeepFake detectors by benchmarking their performance in white-box, gray-box, and black-box settings. They found that, when tested ``in the wild'', all 51 detectors exhibit less than 70\% AUC. This highlights a significant gap between current research efforts and real-world performance, indicating that despite substantial progress, there remains considerable room for improvement—especially in practical settings.
In a recent work \cite{DBLP:conf/cvpr/WangBZWL23}, the authors propose an approach that combines spatial and temporal forgery clues, with the goal of creating a more robust detector. Specifically, starting from a 3D spatiotemporal CNN, they split the training parameters into two groups, spatial- and temporal-related. Then, for each training iteration, they freeze one of the groups while updating only the weights of the remaining group, alternating between the two. This approach results in a much more robust detector that outperforms previous state-of-the-art methods \cite{DBLP:conf/iccv/ZhengB0ZW21,DBLP:conf/cvpr/HaliassosMPP22,DBLP:conf/cvpr/ShioharaY22,DBLP:conf/cvpr/LiBZYCWG20,DBLP:conf/cvpr/ChenZSLW22} on unseen datasets.

\textbf{Contrastive Methods.} Contrastive learning has become a widely used approach in deep learning, particularly for computer vision tasks. A notable example is \cite{DBLP:conf/icml/ChenK0H20}, which contrasts different augmented views of the same input to maximize representation agreement. This enabled the model to outperform previous state-of-the-art results on ImageNet using only a linear classifier trained on the learned embeddings.
In the context of DeepFake detection, \cite{DBLP:conf/ih/ShenZQ22} focused on identity-related information by using contrastive loss to distinguish between real and fake identities. Specifically, the employed sampling method allows contrasting only samples from the same identity, with the aim of making the model pay more attention to the distinctions between real and fake, rather than focusing on identity-related features. \cite{DBLP:conf/aaai/SunYCDLJ22} proposed a dual contrastive learning strategy for uncovering common forgery clues. They employed both inter-instance and intra-instance contrastive learning—pulling together different views of the same sample while reinforcing feature homogeneity. Furthermore, \cite{DBLP:conf/iwbf/KumarBV20} found that using triplet loss in combination with FaceNet \cite{DBLP:conf/cvpr/SchroffKP15} embeddings outperformed other approaches on the highest compression rate on FaceForensics++ \cite{DBLP:conf/iccv/RosslerCVRTN19}.

\textbf{DeepFake Attribution.} Compared to the vast body of research on binary DeepFake detection, studies on DeepFake attribution remain limited. In \cite{DBLP:conf/mmsp/JainKM21}, the authors train an attribution model using a Siamese network with Triplet loss, and use the learned representations to train a neural network classifier. Their results indicate that contrastive training can enhance cross-dataset generalization in DeepFake attribution. Another study \cite{DBLP:conf/icip/JiaLL22}, proposed using CNN extracted features in conjunction with the \textit{CBAM} 2D spatial attention module \cite{DBLP:conf/eccv/WooPLK18} for isolating important spatial artifacts. Moreover, the authors used a Temporal Attention Map to aggregate frame-level features and further isolate the fingerprint of each manipulation method. They trained and tested their approach on their proposed dataset (DFDM), consisting of five faceswap manipulation methods. Their findings demonstrate that their method performs better than other existing GAN attribution methods \cite{DBLP:conf/mipr/MarraGVP19,DBLP:journals/pami/AsnaniYHL23}. Finally, \cite{DBLP:journals/tcsv/LiNYFZ23} presents a multi-scale approach that disentangles manipulation artifacts from irrelevant features using adversarial training and pixel-level supervision. When tested for attribution the proposed method exhibits superior performance compared to CNNs. However, none of the above works examined whether their attribution models maintain their performance when tested in cross-dataset settings, i.e. whether DeepFake attribution models generalize to known manipulation methods from different datasets.

\section{Methodology}

\subsection{\textbf{RQ1: Cross-Dataset Generalization}}
\label{rq1}
To address RQ1, we investigate how training for attribution impacts cross-dataset generalization. We use four model architectures known for their robust performance in image and video classification, adapting each for both binary DeepFake detection and multi-class attribution. Models are trained on FF++ (c23 compression) and evaluated on FF++, CelebDF, and DFDC. 
Since the target datasets include similar but not identical sets of manipulation methods, a direct comparison between multi-class and binary models is not straightforward. Specifically, multi-class attribution models are trained to recognize manipulation types, but when applied to datasets containing unseen or non-overlapping manipulations, their output classes may not align with the available ground truth. To address this issue and ensure fair, consistent evaluation across datasets using the evaluation metrics defined in Section \ref{sec:metrics}, we convert both the multi-class predictions and ground truth labels to a binary format—labeling all manipulated content as \texttt{fake} and authentic content as \texttt{real}.

Given a dataset \( \mathcal{D} \) with \( N \) manipulation methods, represented as a set \(  \mathcal{M} = \{m_1, \ldots, m_N\} \), the labels of this dataset \( \mathcal{D} \) are defined in the set \( \{0, 1, \ldots, N\} \), where \( 0 \) represents the \texttt{real} class and the numbers \( 1 \) to \( N \) correspond to the known manipulation methods. To convert these multi-class labels \( l_{\text{mc}} \) into a binary format, we define the corresponding binary label \( l_{\text{bin}} \) as:

\begin{equation}
l_{\text{bin}} = 
\begin{cases} 
0 & \text{if } l_{\text{mc}} = 0, \\
1 & \text{if } l_{\text{mc}} > 0.
\end{cases}
\label{eq:lbin}
\end{equation}

To calculate metrics such as AUC that require prediction scores, we define a procedure for converting multi-class prediction scores to binary. Consider a multi-class classifier \( F(X;\theta) \), where \( X \) is a sample image with dimensions \(H \times W \ \times 3\) and \(\theta\) the classifier's parameters. The output, \( \mathbf{p}_{\text{mc}} \), is a softmax vector of size \( N+1 \), with each element representing the confidence score for a given class, indicating the relative likelihood of each class. Let \( j = \underset{i}{\arg\max} \ \mathbf{p}_{\text{mc}}[i] \) denote the index of the maximum probability, and \( p_{\text{max}} = \mathbf{p}_{\text{mc}}[j] \) the corresponding value. The binary prediction \( p_{\text{bin}} \) is defined as:

\begin{equation}
p_{\text{bin}} = 
\begin{cases} 
p_{\text{max}} & \text{if } j = 0 \text{ and } p_{\text{max}} < 0.5, \\
1 - p_{\text{max}} & \text{if } j = 0 \text{ and } p_{\text{max}} \geq 0.5, \\
p_{\text{max}} & \text{if } j \neq 0
\end{cases}
\label{eq:pbin}
\end{equation}

Based on the binary label $l_{\text{bin}}$ and binary prediction $p_{\text{bin}}$ of the multi-class models, we calculate the evaluation metrics described in section \ref{sec:metrics}. We then compare these results with the corresponding metrics of the binary models. 

\subsection{\textbf{RQ2: Intra-Manipulation Performance}}
\label{rq2}
For RQ2, we shift our focus to evaluating same-manipulation performance across datasets. Specifically, we assess model performance on test samples that involve manipulation methods seen during training but originating from different datasets. To address RQ2, we identify pairs of manipulation methods that are shared across datasets, enabling a controlled comparison of generalization in view of data distribution shifts. 

Specifically, we define \( \mathcal{D} = \{\mathcal{D}_0, \mathcal{D}_1, \mathcal{D}_2, \ldots, \mathcal{D}_n\}\) as a set of datasets and \(\mathcal{M} = \{m_0, m_1, m_2, \ldots, m_N\}\) as the set of all unique manipulations across these datasets. Each dataset \( \mathcal{D}_i \in  \mathcal{D}\) contains a subset of manipulations \( \mathcal{M}_{\mathcal{D}_i} \subseteq \mathcal{M} \). We identify ordered pairs \( (\mathcal{D}_i,  \mathcal{D}_j)\) for \( \mathcal{D}_i,  \mathcal{D}_j \in  \mathcal{D}\) and \(i \neq j\), and determine every unique manipulation \(m_k\) that they have in common:

\begin{equation}
\{(\mathcal{D}_i, \mathcal{D}_j, m_k) \mid m_k \in M_{\mathcal{D}_i} \cap M_{\mathcal{D}_j}, \, i \neq j\}
\end{equation}

This process leads to the creation of specific dataset pairs for each common manipulation method \( m_k \), which we treat as separate training and testing instances. For each backbone \( b_i \) listed in Table \ref{tab:models}, and for every identified triplet \((\mathcal{D}_i, \mathcal{D}_j, m_k)\) we follow the training and evaluation protocol described below:

\begin{itemize}
    \item \textbf{Training:} Train the backbone model \( b_i \) on the full training dataset \( \mathcal{D}_i \) using multi-class classification, where manipulation \( m_k \) is included alongside other manipulation types.
    \item \textbf{Testing:} Evaluate the model, trained on \(\mathcal{D}_i\) with backbone \(b_i\), on the manipulation \(m_k\) present in dataset \(\mathcal{D}_j\).
\end{itemize}

This approach allows us to systematically investigate feature generalization and model robustness, offering an evaluation framework that specifically measures performance on known manipulation methods across different datasets.

\subsection{\textbf{RQ3: Impact of Contrastive Methods on Attribution}}
For RQ3, we enhance the attribution models used in RQ1 by incorporating contrastive loss functions, including Triplet loss \cite{DBLP:journals/jmlr/WeinbergerS09}, NT-Xent loss \cite{DBLP:conf/icml/ChenK0H20,DBLP:conf/cvpr/He0WXG20,DBLP:journals/corr/abs-1807-03748}, and SupCon loss \cite{DBLP:conf/nips/KhoslaTWSTIMLK20}. 
Contrastive learning aims to structure the embedding space such that samples with similar characteristics (e.g., generated by the same manipulation method) are pulled closer together, while dissimilar samples (e.g., real vs. fake, or fakes from different sources) are pushed apart. This is particularly beneficial for DeepFake attribution, where subtle artifacts introduced by different manipulation methods may not be easily captured by traditional classification loss alone, leading to manipulation-aware representations.
Comparative analysis between vanilla and contrastive-enhanced models is conducted to determine the effectiveness of contrastive methods in improving attribution performance. Next, we present an overview of the used contrastive loss functions used.

\textbf{Triplet loss \cite{DBLP:journals/jmlr/WeinbergerS09}}: This seeks to minimize the distance between the anchor \(a\) and the positive sample \(p\), while maximizing the distance between the anchor and the negative sample \(n\). It is designed to ensure that \(a\) is closer to \(p\) than to \(n\) by at least a margin \(m\). 

\begin{equation}
L(a, p, n) = \max(0, \|f(a) - f(p)\|^2 - \|f(a) - f(n)\|^2 + m)
\label{eq:triplet_loss}
\end{equation}

Here, \(f(x)\) is the feature representation of $x$, and \(a\), \(p\), and \(n\) represent the anchor, positive, and negative samples, respectively. Triplet loss is one of the most popular contrastive losses, in both  supervised and self-supervised settings in computer vision. However, its power is limited by the fact that it utilizes only one positive and one negative sample. Retrieving a large number of hard triplets is typically required to achieve good performance \cite{DBLP:journals/jmlr/WeinbergerS09,DBLP:conf/cvpr/ChopraHL05,DBLP:conf/cvpr/SchroffKP15}.

\textbf{NT-Xent}: The normalized temperature-scaled cross entropy, improves the contrastive power of triplet loss by using multiple negative samples instead of one. It is a generalization of N-Pair loss \cite{DBLP:conf/nips/Sohn16} that has been shown to outperform Triplet loss \cite{DBLP:conf/nips/Sohn16,DBLP:conf/cvpr/WuXYL18,DBLP:journals/corr/abs-1807-03748}.

\begin{equation}
L(i, j) = -\log \left(\frac{\exp(\cos(f(x_i), f(x_j)) / \tau)}{\sum_{k=1}^{2N} \mathbf{1}_{k \neq i} \exp(\cos(f(x_i), f(x_k)) / \tau)}\right)
\label{eq:ntxent_loss}
\end{equation}

In the above definition, \(\cos\) represents the cosine similarity between  feature representations, \(\tau\) denotes a temperature parameter, \(x_i\) and \(x_j\) are positive pairs, and $N$ is the batch size.

\textbf{SupCon \cite{DBLP:conf/nips/KhoslaTWSTIMLK20}}: Supervised contrastive loss can be seen as a generalization of Triplet and N-Pair losses. It directly incorporates label information, enabling supervised contrastive learning.
\begin{equation}
L = \sum_{i=1}^N \frac{-1}{|P_i|} \sum_{p \in P(i)} \log \left(\frac{\exp( \cos(f(x_i), f(x_p)) / \tau)}{\sum_{a \in A_i} \exp(\cos(f(x_i), f(x_a)) / \tau)}\right)
\label{eq:supcon_loss}
\end{equation}

Here, \(P_i\) is the set of indices of all positives in the batch for the anchor, \(A_i\) includes all other examples in the batch except \(i\), and \(\tau\) is the temperature parameter.

\section{Experimental Setup}
This study employs a comparative and evaluative research design to address the outlined research questions regarding the performance of DeepFake detection and attribution models under various conditions. The methodology is structured into three main components corresponding to each research question (RQ).

\subsection{Datasets \& Preprocessing}

We utilize the following established DeepFake datasets also summarized in Table \ref{tab:deepfake_datasets}:

\begin{itemize}
\item \textbf{FaceForensics++ (FF++) \cite{DBLP:conf/iccv/RosslerCVRTN19}} includes 1000 real and 1000 manipulated videos per manipulation type: FaceSwap, DeepFakes, Face2Face, and NeuralTextures. An updated version includes also FaceShifter. All videos are sourced from YouTube and available in three compression levels: c0 (none), c23 (low), and c40 (high). We use  c23 in this work.
\item \textbf{CelebDF-V2 (CelebDF) 
 \cite{DBLP:conf/cvpr/LiYSQL20}} is considered a second generation dataset, due to its good quality of FaceSwap DeepFakes. The manipulation method used is the same as in FF++ DeepFakes but with multiple visual quality improvements. It contains 590 real and 5,639 fake videos, sourced from celebrity interviews on YouTube.
\item \textbf{FakeAVCeleb \cite{DBLP:conf/nips/KhalidTKW21}} is an audio-visual DeepFake dataset containing 500 real and 19,500 fake videos, with real videos originating from the VoxCeleb2 \cite{DBLP:conf/interspeech/ChungNZ18} dataset. The fake videos were manipulated with four methods: FaceSwap, FSGAN, Wav2Lip, and SV2TTS. The present work makes use of the video manipulations only. 
\item \textbf{DFDC \cite{DBLP:journals/corr/abs-2006-07397}} is one of the largest publicly available collection of DeepFake videos. Although the dataset lacks explicit manipulation labels, it primarily consists of FaceSwap manipulations. It consists of $\approx$ 24K real and $\approx$ 105K fake videos.
\item \textbf{ForgeryNet \cite{DBLP:conf/cvpr/HeGCZYSSS021}} is one of the largest publicly available DeepFake datasets. It contains $\approx$ 100K real and $\approx$ 120K fake videos, generated from 8 manipulation methods.
\item \textbf{DFPlatter} \cite{DBLP:conf/cvpr/NarayanATMV023} is the most recent dataset, comprising over 133K videos featuring subjects primarily of Indian ethnicity. It includes three manipulations (FaceSwap, FSGAN, and FaceShifter), and is divided into three sets. In this study, we use for simplicity only the single subject videos of Set A, although the same methodology can be applied on sets B and C featuring multiple individuals per video.

\end{itemize}

For each video in the datasets, we sample one frame per second using \textit{FFMPEG} \cite{tomar2006converting}, as the detection models employed are frame-based. Subsequently, faces are detected and extracted using the RetinaFace \cite{DBLP:conf/cvpr/DengGVKZ20} face detector. Regarding train/test splitting, we followed the original splits but applied them at the frame level. This means that training frames were extracted from videos in the original training split of a dataset, and test frames from the test split.

\begin{table}[h]
\centering
\begin{tabular}{l|c|c|c}
\hline
\textbf{Dataset} & \textbf{Real} & \textbf{Fake} & \textbf{\# Manip.} \\
\hline
FaceForensics++ \cite{DBLP:conf/iccv/RosslerCVRTN19} & 1000 & 1000/type & 5 \\
CelebDF-V2 \cite{DBLP:conf/cvpr/LiYSQL20} & 590 & 5639 & 1 \\
FakeAVCeleb \cite{DBLP:conf/nips/KhalidTKW21} & 500 & 19.5K & 4 \\
DFDC \cite{DBLP:journals/corr/abs-1910-08854}& $\sim$24K & $ \sim$105K & 1 \\
ForgeryNet \cite{DBLP:conf/cvpr/HeGCZYSSS021} & $\sim$100K & $ \sim$120K & 8 \\
DFPlatter \cite{DBLP:conf/cvpr/NarayanATMV023}  & 764 & $\sim$133K & 3 \\
\hline
\end{tabular}
\caption{Overview of Deepfake Datasets}
\label{tab:deepfake_datasets}
\end{table}

\subsection{Selected Models}

For our experiments, we selected five models: two based on convolutional neural networks (CNNs), two based on transformer architectures (ViT), and a hybrid model that combines CNNs and ViTs, specifically designed for DeepFake detection \cite{coccomini2022combining}. Table~\ref{tab:models} summarizes the input resolutions and number of parameters for comparison.

\begin{table}[tb]
    \centering
    \small
    \caption{Parameters of Selected Models}
    \label{tab:models}
    \resizebox{\columnwidth}{!}{%
    \begin{tabular}{@{}lccc@{}}
        \toprule
        Architecture & Model & Input Size & Parameters \\
        \midrule
        \multirow{2}{*}{CNN} 
            & EfficientNetV2 B0 & \(224 \times 224 \times 3\) & 5.87M \\
            & ConvNextV2 Tiny & \(384 \times 384 \times 3\) & 27.9M \\
        \midrule
        \multirow{2}{*}{Transformer} 
            & PyramidNetV2 B0& \(224 \times 224 \times 3\) & 3.41M \\
            & SwinV2 Tiny & \(256 \times 256 \times 3\) & 27.58M \\
        \midrule
        CNN + ViT 
            & Efficient ViT & \(224 \times 224 \times 3\) & 109M \\
        \bottomrule
    \end{tabular}%
    }
\end{table}

\begin{itemize}

\item \textbf{EfficientNetV2} \cite{DBLP:conf/icml/TanL21}: A convolutional model family designed via Neural Architecture Search and a new scaling method, offering improved speed and parameter efficiency over the original EfficientNet \cite{DBLP:conf/icml/TanL19}. 
EfficientNet variants have been widely used for DeepFake detection \cite{DBLP:conf/cvpr/ShioharaY22,DBLP:conf/cvpr/ZhaoZ0WZY21,DBLP:journals/corr/abs-1910-08854,ferrer2020deepfake,DBLP:conf/cvpr/HuangWYA00Y23}, affirming its use in our study.

\item  \textbf{ConvNextV2} \cite{DBLP:conf/cvpr/WooDHC0KX23}: is an improvement of the previous ConvNext \cite{DBLP:conf/cvpr/0003MWFDX22}, which modernized a vanilla ResNet towards the design of Vision Transformers. In contrast to the first version, ConvNextV2 models are pre-trained using Fully Convolutional Masked AutoEncoders (FCMAE), utilize Sparse Convolutions, and include a Global Response Normalization module that improves inter-channel feature competition. It is recognized as a cutting-edge backbone, demonstrating exceptional performance in downstream tasks \cite{DBLP:conf/nips/GoldblumSNSPSCI23}.

\item \textbf{Pyramid Vision Transformer V2 (PVT-V2)} \cite{DBLP:journals/cvm/WangXLFSLLLS22}: A hierarchical vision transformer that uses a pyramid structure to reduce computational cost compared to standard Vision Transformers (ViTs) \cite{DBLP:conf/iclr/DosovitskiyB0WZ21}. Its design supports fine-grained inputs and smaller patches, producing multi-scale feature maps. PVT-V2 simplifies the architecture further by replacing positional embeddings with zero-padding and overlapping patch embeddings.

\item \textbf{SwinV2} \cite{DBLP:conf/cvpr/Liu0LYXWN000WG22}: A general-purpose vision transformer that uses a shifted-window mechanism to limit self-attention to local regions while enabling cross-window connections. It generates hierarchical feature maps with linear complexity relative to image size. SwinV2 introduces architectural improvements for training stability and supports self-supervised pre-training via SimMIM \cite{xie2022simmim}. Its effectiveness across vision tasks has been empirically validated \cite{DBLP:conf/nips/GoldblumSNSPSCI23}.

\item \textbf{Efficient ViT} \cite{coccomini2022combining}: This  combines a convolutional backbone with a Transformer encoder, based on the structure of a Vision Transformer. Specifically, it employs EfficientNet-B0 as the convolutional feature extractor to process input faces. The output consists of visual features corresponding to 7×7 pixel chunks of the input. These features are then linearly projected and fed into a Vision Transformer. A CLS token is then used as the learned representation for binary or multiclass classification.

\end{itemize}

While there are other state-of-the-art models that also combine convolution with transformers using distillation for training \cite{heo2021deepfake}, we do not include such models in this study, as their performance heavily relies on the distillation approach. In this work, we focus on investigating the effect of incorporating contrastive loss alongside a multiclass classification loss.

\subsubsection{\textbf{Implementation Details}}
The first four models are implemented using the PyTorch Image Models collection\footnote{\url{https://huggingface.co/timm}} and PyTorch Lightning\footnote{\url{https://lightning.ai/docs/pytorch/stable/}}, while for Efficient ViT we used the implementation provided by the authors\footnote{\url{https://tinyurl.com/cnn-vit-dfd}}. Training was performed on NVIDIA GPUs. During training, we applied data augmentation techniques such as random cropping, AugMix \cite{DBLP:conf/iclr/HendrycksMCZGL20}, and horizontal flipping.

\subsubsection{\textbf{Contrastive Loss Implementation Details}} 
To apply NT-Xent and SupCon losses to DeepFake attribution, we adopt the SimCLR framework \cite{DBLP:conf/icml/ChenK0H20} and incorporate a trainable projection head on top of the encoder. This projection head is a multi-layer perceptron (MLP) with one hidden layer and ReLU activation, which maps the high-dimensional encoder output into a lower-dimensional latent space, reducing the encoder’s output by a factor of 16. Contrastive loss is computed on the outputs of this projection head, rather than directly on the encoder features. At test time, the projection head is removed, and only the encoder is used. Additionally, for each sample in a batch of size \( N \), we generate two augmented views forming a positive pair, resulting in a batch size of \( 2N \).

\subsection{\textbf{Evaluation Metrics}}\label{sec:metrics}
Performance is assessed using AUC, Equal-Error-Rate, and Balanced Accuracy to provide a comprehensive view across different conditions.

\begin{itemize}

\item  \textbf{AUC}: Evaluates a model's ability to distinguish between classes. It represents the probability that a randomly selected positive instance is ranked higher than a randomly selected negative one. AUC is especially useful for comparing models across different decision thresholds.

\item  \textbf{Equal-Error Rate (EER)}: The point where the false acceptance rate equals the false rejection rate, commonly used in biometric systems. In DeepFake detection, lower EER values indicate better balance between false positives and false negatives. 

\item  \textbf{Balanced Accuracy (BA)}: Measures average recall across classes, addressing class imbalance by equally weighting each class, which is useful in imbalanced DeepFake datasets.
\end{itemize}

\section{Results and Discussion}

\begin{figure}[!htbp]
    \centering
    \includegraphics[width=1.0\columnwidth, alt={Comparison of AUC, Balanced Accuracy (BA), and Equal Error Rate (EER) for Binary and Multiclass models trained on Faceforensics++.}]{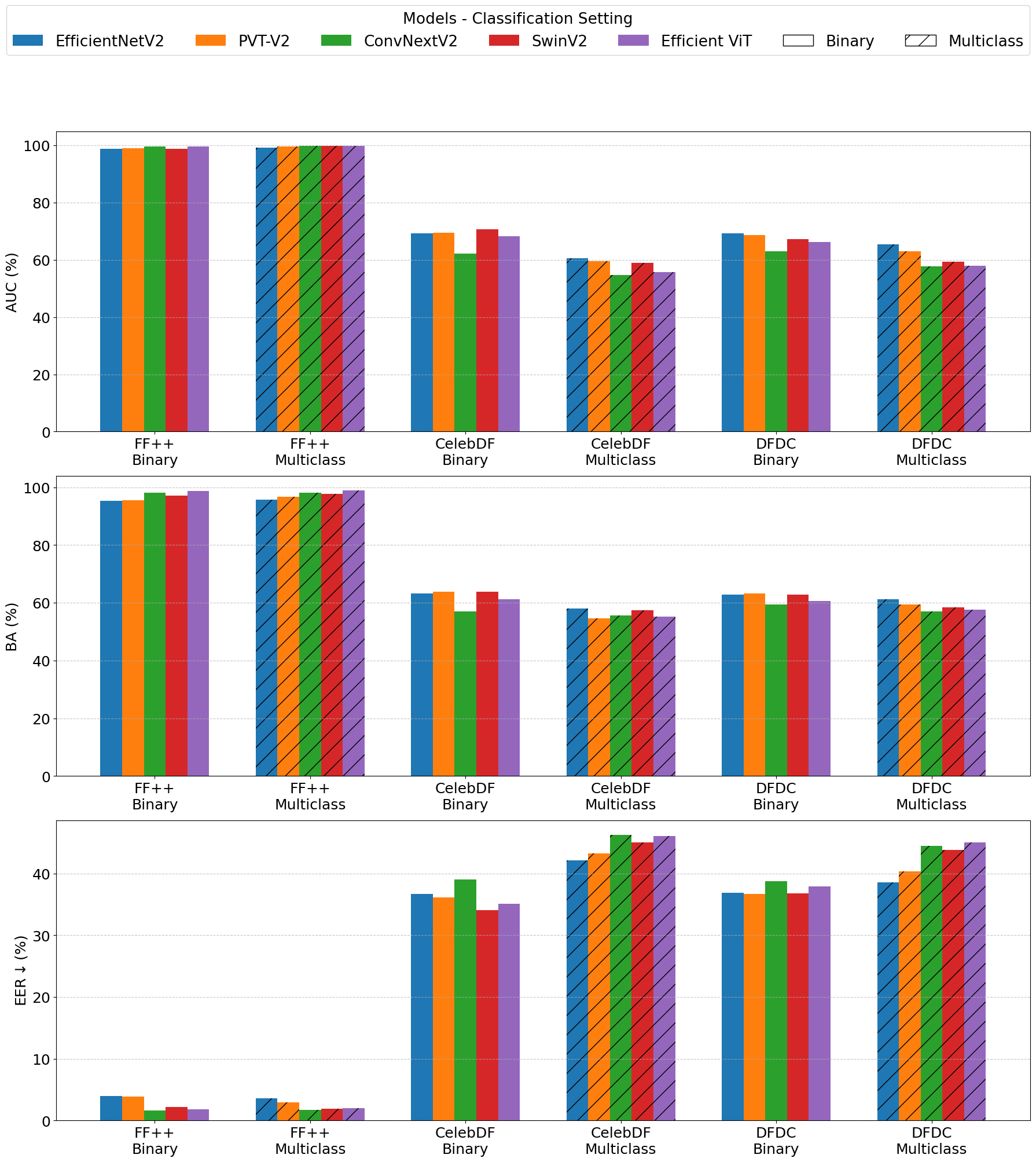}
    \caption{Comparison of AUC, Balanced Accuracy (BA) and Equal Error Rate (EER) for Binary and Multiclass models trained on FaceForensics++ (FF++). Evaluation is performed on FaceForensics++, CelebDF, and DFDC.}
    \label{fig:auc_ba_eer_ff}
\end{figure}

Figure~\ref{fig:auc_ba_eer_ff} presents the experimental results for RQ1, where detectors are trained on FaceForensics++ and evaluated on FaceForensics++, CelebDF, and DFDC. A key observation is that within-dataset performance is significantly higher than cross-dataset performance. In the case of FaceForensics++, all evaluation metrics are near-optimal, with AUC scores approaching 100\% across all models. Moreover, the results indicate that binary classification models consistently demonstrate better cross-dataset generalization compared to multi-class DeepFake detectors. Notably, within the same dataset (i.e., both training and testing are performed on FaceForensics++), multi-class detectors achieve slightly higher AUC and Balanced Accuracy (BA) than binary models. This suggests that multi-class detectors are capable of learning features useful for manipulation attribution, but these appear to be dataset-specific and do not transfer well across datasets. The superior cross-dataset performance of binary models may be attributed to the increased diversity within each class, allowing models to learn more general representations.

\begin{figure}[!htbp]
    \centering
    \includegraphics[width=1.0\columnwidth]{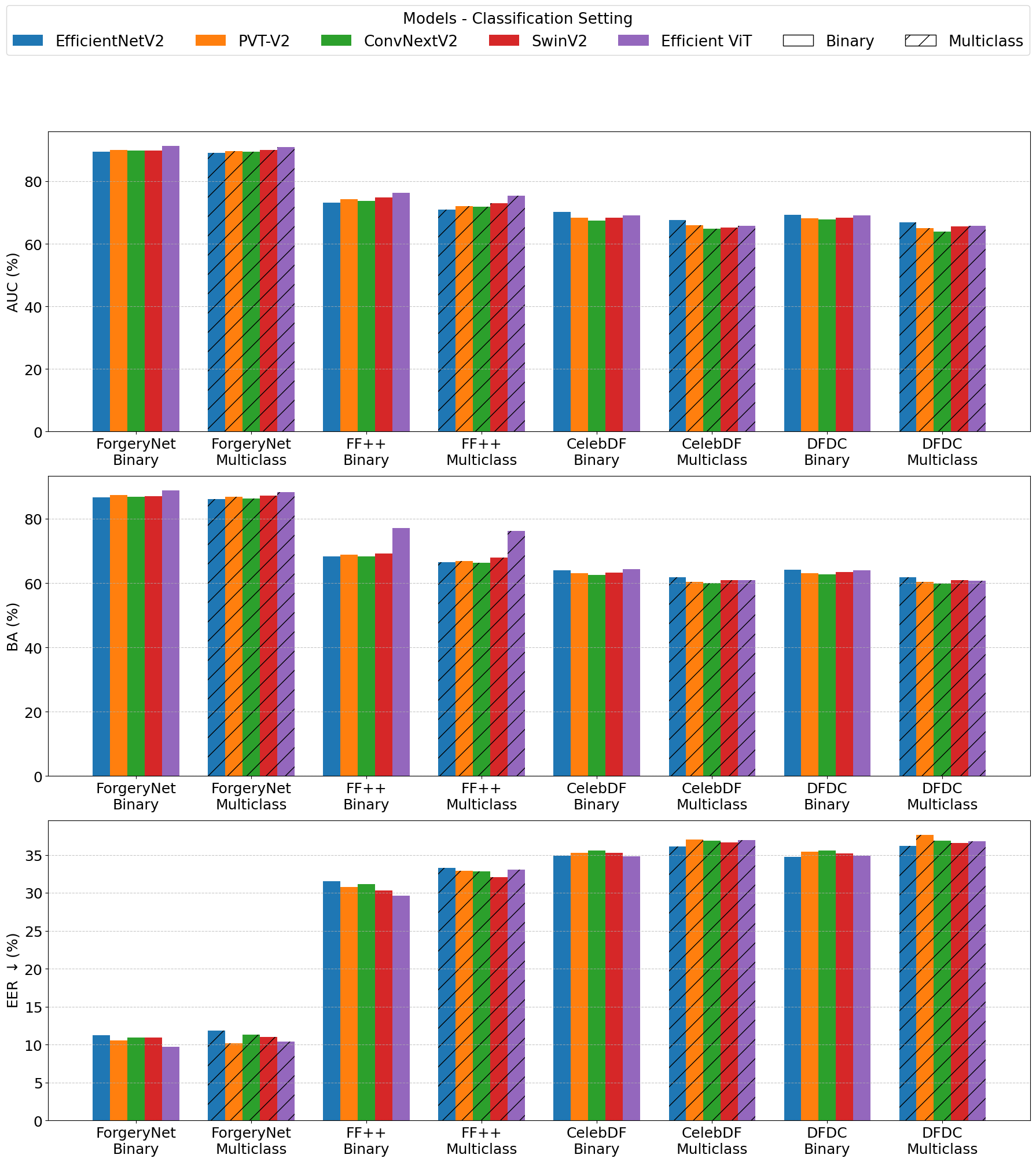}
    \caption{Comparison of AUC, Balanced Accuracy (BA) and Equal Error Rate (EER) for Binary and Multiclass models trained on ForgeryNet. Evaluation is performed on ForgeryNet, FaceForensics++ (FF++), CelebDF, and DFDC.}
    \label{fig:auc_ba_eer_forgerynet}
\end{figure}

To validate these findings, we conducted an additional experiment where detectors were trained on ForgeryNet and evaluated on ForgeryNet, CelebDF, FaceForensics++, and DFDC. As shown in Figure~\ref{fig:auc_ba_eer_forgerynet}, within-dataset performance (i.e., trained and tested on ForgeryNet) remains high, but is lower than the within-dataset performance achieved when training and testing on FaceForensics++. This can be attributed to the more challenging nature of ForgeryNet, especially when compared to FaceForensics++, which is widely considered an ``easy'' dataset, as it does not reflect the variety and realism of more recent synthetic media. In addition, the multiclass setting achieves results comparable to the binary one on ForgeryNet, highlighting the feasibility of attribution models. However, as shown in the other datasets of Figure~\ref{fig:auc_ba_eer_forgerynet}, all detectors exhibit limited generalization in the cross-dataset setting, with accuracy dropping from around 90\% to 76\% in the best case and as low as 67\% in the worst case. Moreover, binary classifiers consistently outperform their multi-class counterparts under these conditions. However, it is worth noting that the performance gap between binary and multi-class detectors is smaller than in the previous setting. This suggests that larger and more recent datasets like ForgeryNet may improve generalization performance across different datasets.

Regarding RQ2, as illustrated in Tables~\ref{tab:rq2_deepfakes_only}, \ref{tab:rq2_faceshifter_only}, \ref{tab:rq2_faceswap_only} and \ref{tab:rq2_fsgan_only}, all models struggle to effectively detect seen manipulations in unseen datasets. Even larger models, such as ConvNext and Swin, or the state-of-the-art Efficient ViT experience significant drops in accuracy. Particularly in FaceSwap manipulations (Table \ref{tab:rq2_faceswap_only}) all models are found to completely fail on cross-dataset detection of the same manipulation with accuracy close to 0. Note also that models trained on FaceForensics++ fail to generalize to other datasets across all manipulation types. However, an improvement in generalization performance is noted in models trained on newer or higher-quality datasets like DFPlatter and CelebDF. For instance, in the case of FSGAN manipulations (Table \ref{tab:rq2_fsgan_only}), all five models trained on DFPlatter achieve high accuracy within the dataset and maintain strong performance on FakeAVCeleb, and to a lesser extent, on ForgeryNet. Similarly, models trained on DFPlatter achieve an accuracy of $43-45\%$ in detecting FaceShifter DeepFakes on ForgeryNet, while those trained on FaceForensics++ are only $5-15\%$ accurate. A possible explanation for why models struggle to detect seen manipulations in unseen datasets is the presence of distribution shifts between datasets. These shifts might be attributed to factors such as differences in capture devices, resolution, compression levels, lighting conditions, and overall video quality. Such variations could affect the visual patterns that models rely on, limiting their generalization ability.

\begin{table}[!htbp]
    \small
    \caption{Generalization performance of attribution models on DeepFakes manipulations in terms of accuracy.}
    \label{tab:rq2_deepfakes_only}
    \centering 
    \begin{tabularx}{\columnwidth}{@{}l *{4}{Y}@{}}
        \toprule
        Model & Train\textbackslash Test & FF++ & CelebDF & ForgeryNet \\
        \midrule
        \multirow{3}{*}{EfficientNetV2} 
        & FF++ & \textbf{96.99} & 10.20 & 11.28 \\
        & CelebDF         & 35.82        & \textbf{99.00} & 33.04 \\
        \midrule
        \multirow{3}{*}{PVT-V2} 
        & FF++ & \textbf{97.09} & 26.44 & 15.06 \\
        & CelebDF         & 18.44        & \textbf{98.15} & 20.69 \\
        \midrule
        \multirow{3}{*}{ConvNextV2} 
        & FF++ & \textbf{97.82} & 2.90  & 9.15 \\
        & CelebDF         & 20.95        & \textbf{99.78} & 20.76 \\
        \midrule 
        \multirow{3}{*}{SwinV2} 
        & FF++ & \textbf{97.82} & 7.56  & 5.32 \\
        & CelebDF         & 25.51        & \textbf{98.95} & 14.60 \\
        \midrule 
        \multirow{3}{*}{Efficient ViT} 
        & FF++ & \textbf{98.02} & 18.65 & 14.24 \\
        & CelebDF         & 31.12        & \textbf{99.02} & 27.61 \\
        \bottomrule
    \end{tabularx}
\end{table}

\begin{table}[!htbp]
    \small
    \caption{Generalization performance of attribution models on FaceShifter manipulations in terms of accuracy.}
    \label{tab:rq2_faceshifter_only}
    \centering 
    \begin{tabularx}{\columnwidth}{@{}l *{4}{Y}@{}}
        \toprule
        Model & Train\textbackslash Test & FF++ & ForgeryNet \\
        \midrule
        \multirow{3}{*}{EfficientNetV2} 
        & FF++ & \textbf{97.04} & 14.72 \\
        & DFPlatter       & 33.26        & \textbf{50.40} \\
        \midrule
        \multirow{3}{*}{PVT-V2} 
        & FF++ & \textbf{98.11} & 5.52 \\
        & DFPlatter       & 18.70        & \textbf{43.71} \\
        \midrule
        \multirow{3}{*}{ConvNextV2} 
        & FF++ & \textbf{98.36} & 9.46 \\
        & DFPlatter       & 19.87        & \textbf{41.61} \\
        \midrule 
        \multirow{3}{*}{SwinV2} 
        & FF++ & \textbf{98.08} & 5.04 \\
        & DFPlatter       & 15.14        & \textbf{46.41} \\
        \midrule 
        \multirow{3}{*}{Efficient ViT} 
        & FF++ & 98.08         & 5.04 \\
        & DFPlatter       & \textbf{34.40} & \textbf{49.63} \\
        \bottomrule
    \end{tabularx}
\end{table}

\begin{table}[!htbp]
    \small
    \caption{Generalization performance of attribution models on FaceSwap manipulations in terms of accuracy.}
    \label{tab:rq2_faceswap_only}
    \centering 
    \begin{tabularx}{\columnwidth}{@{}l *{3}{Y}@{}}
        \toprule
        Model & Train\textbackslash Test & FF++ & FakeAVCeleb \\
        \midrule
        \multirow{3}{*}{EfficientNetV2} 
        & FF++ & \textbf{95.47} & 0.00 \\
        & FakeAVCeleb & 0.86 & \textbf{95.64} \\
        \midrule
        \multirow{3}{*}{PVT-V2} 
        & FF++ & \textbf{97.32} & 0.00 \\
        & FakeAVCeleb & 0.54 & \textbf{93.67} \\
        \midrule
        \multirow{3}{*}{ConvNextV2} 
        & FF++ & \textbf{96.58} & 0.00 \\
        & FakeAVCeleb & 0.08 & \textbf{96.48} \\
        \midrule 
        \multirow{3}{*}{SwinV2} 
        & FF++ & \textbf{96.46} & 0.00 \\
        & FakeAVCeleb & 0.08 & \textbf{96.91} \\
        \midrule 
        \multirow{3}{*}{Efficient ViT} 
        & FF++ & \textbf{97.27} & 0.00 \\
        & FakeAVCeleb & 1.72 & \textbf{95.98} \\
        \bottomrule
    \end{tabularx}
\end{table}

\begin{table}[!htbp]
    \small
    \caption{Generalization performance of attribution models on FSGAN manipulations in terms of accuracy.}
    \label{tab:rq2_fsgan_only}
    \centering 
    \begin{tabularx}{\columnwidth}{@{}l *{4}{Y}@{}}
        \toprule
        Model & Train\textbackslash Test & DFPlatter & FakeAVCeleb & ForgeryNet \\
        \midrule
        \multirow{3}{*}{EfficientNetV2} 
        & FakeAVCeleb     & 38.52        & \textbf{98.20} & 38.16 \\
        & DFPlatter      & \textbf{99.59} & 98.12        & 69.42 \\
        \midrule
        \multirow{3}{*}{PVT-V2} 
        & FakeAVCeleb     & 17.80        & \textbf{97.11} & 17.02 \\
        & DFPlatter      & \textbf{98.77} & 99.46        & 62.66 \\
        \midrule
        \multirow{3}{*}{ConvNextV2} 
        & FakeAVCeleb     & 22.81        & \textbf{99.72} & 12.28 \\
        & DFPlatter      & \textbf{99.63} & 99.77        & 66.39 \\
        \midrule 
        \multirow{3}{*}{SwinV2} 
        & FakeAVCeleb     & 12.91        & \textbf{99.79} & 10.93 \\
        & DFPlatter      & \textbf{99.90} & 100.00       & 69.24 \\
        \midrule 
        \multirow{3}{*}{Efficient ViT} 
        & FakeAVCeleb     & 38.01        & \textbf{99.32} & 37.96 \\
        & DFPlatter      & \textbf{99.61} & 98.12        & 69.41 \\
        \bottomrule
    \end{tabularx}
\end{table}

\begin{table}[htbp]
    \small
    \caption{Comparison of CNN, ViT and CNN+ViT architectures across different manipulations for all results, within-dataset only, and cross-dataset only evaluations (accuracy \%).}
    \label{tab:cnn_vit_comparison}
    \begin{tabularx}{\columnwidth}{@{}l *{5}{Y}@{}}
        \toprule
        \multirow{2}{*}{Architecture} & \multicolumn{4}{c}{Manipulation} & \multirow{2}{*}{Average} \\
        \cmidrule(lr){2-5}
         & \rotatebox{20}{DeepFakes} & \rotatebox{20}{FaceShifter} & \rotatebox{20}{FaceSwap} & \rotatebox{20}{FSGAN} & \\
        \midrule
        \multicolumn{6}{c}{All} \\
        \midrule
        CNN & 44.81 & \textbf{47.51} & 48.14 & \textbf{69.27} & \textbf{52.43} \\
        ViT & 43.80 & 44.36 & 48.12 & 65.47 & 50.44 \\
        CNN + ViT & \textbf{44.91} & 46.61 & \textbf{50.01} & 68.78 & 51.24 \\
        \midrule
        \multicolumn{6}{c}{Intra-dataset} \\
        \midrule
        CNN & \textbf{98.40} & 98.78 & 96.04 & \textbf{99.23} & \textbf{98.11} \\
        ViT & 98.00 & \textbf{98.97} & 96.09 & 98.89 & 97.99 \\
        CNN + ViT & 98.11 & \textbf{98.97} & \textbf{96.31} & 99.07 & 98.07 \\
        \midrule
        \multicolumn{6}{c}{Cross-dataset} \\
        \midrule
        CNN & \textbf{18.01} & \textbf{21.88} & 0.24 & \textbf{54.28} & \textbf{23.60} \\
        ViT & 16.70 & 17.05 & 0.15 & 48.75 & 20.67 \\
        CNN + ViT & 17.90 & 21.76 & \textbf{0.26} & 53.84 & 22.97 \\
        \bottomrule
    \end{tabularx}
\end{table}

\begin{table*}[!htbp]
\small
  \caption{Comparison between vanilla and contrastive attribution models for RQ3. All models are trained on FF++.
  Legend: B: Baseline, vanilla multiclass training; T-H: Triplet with hard mining; T-HS: Triplet with hard positive and semihard negative mining; SC: Supervised Contrastive loss with 2 views and projection head; NT: NT-Xent loss with 2 views and projection head.}
  \label{tab:rq3}
  \begin{tabularx}{\textwidth}{@{}l *{9}{Y}@{}}
    \toprule
    Setting\textbackslash Dataset & \multicolumn{3}{c}{FaceForensics++} & \multicolumn{3}{c}{CelebDF} & \multicolumn{3}{c}{DFDC} \\
    \cmidrule(lr){2-4} \cmidrule(lr){5-7} \cmidrule(lr){8-10}
    & AUC (\%) & BA (\%) & EER $\downarrow$ (\%) & AUC (\%) & BA (\%) & EER $\downarrow$ (\%) & AUC (\%) & BA (\%) & EER $\downarrow$ (\%) \\
    \midrule
    \multicolumn{10}{c}{EfficientNetV2} \\
    \midrule
    B   & \textbf{99.20} & \textbf{95.66} & \textbf{3.64} & \underline{60.56} & \textbf{57.95} & \underline{42.08} & \textbf{65.47} & \textbf{61.26} & \textbf{38.52} \\
    T-H & 98.08 & 92.43 & \underline{6.86} & \textbf{61.55} & \underline{57.40} & \textbf{41.07} & \underline{64.29} & \underline{59.16} & \underline{39.91} \\
    T-HS& 84.75 & 68.70 & 22.15 & 58.74 & 55.37 & 44.01 & 62.41 & 57.32 & 41.10 \\
    NT  & \underline{98.73} & \underline{95.57} & 4.41 & 60.04 & 54.84 & 42.25 & 64.02 & 58.89 & 40.29 \\
    SC  & 97.70 & 91.68 & 7.93 & 58.31 & 54.02 & 43.05 & 61.85 & 58.48 & 41.73 \\
    \midrule
    \multicolumn{10}{c}{PVT-V2} \\
    \midrule
    B   & \textbf{99.47} & \textbf{96.69} & \textbf{2.93} & \underline{59.53} & \underline{54.59} & \underline{43.27} & \textbf{63.02} & \textbf{59.33} & \textbf{40.36} \\
    T-H & 50.80 & 50.00 & 50.00 & 51.20 & 50.00 & 50.00 & 51.20 & 50.00 & 50.00 \\
    T-HS& 96.99 & 90.27 & 8.09 & 57.33 & 53.14 & 46.49 & \underline{63.63} & \underline{59.10} & \underline{40.58} \\
    NT  & \underline{99.23} & \underline{96.16} & \underline{3.51} & \textbf{59.56} & 54.49 & \textbf{42.28} & 62.47 & 58.19 & 40.77 \\
    SC  & 97.65 & 91.74 & 7.69 & 56.61 & \textbf{54.76} & 46.11 & 56.61 & 54.76 & 46.11 \\
    \midrule
    \multicolumn{10}{c}{ConvNextV2} \\
    \midrule
    B   & \textbf{99.83} & \textbf{98.17} & \underline{1.71} & \textbf{54.70} & \textbf{55.56} & 46.26 & 57.77 & 57.03 & 44.41 \\
    T-H & 99.62 & \underline{98.04} & \textbf{1.55} & \underline{52.55} & 53.20 & \underline{49.08} & 58.89 & 58.54 & 43.53 \\
    T-HS& \underline{99.82} & 97.91 & 1.91 & 50.55 & \underline{53.28} & 48.37 & 60.14 & \underline{58.74} & 42.70 \\
    NT  & 99.56 & 97.58 & 2.20 & 46.44 & 51.61 & 48.28 & \underline{60.36} & 58.06 & \underline{42.39} \\
    SC  & 99.44 & 95.91 & 3.30 & 50.36 & 50.94 & \textbf{49.82} & \textbf{61.21} & \textbf{58.93} & \textbf{41.72} \\
    \midrule
    \multicolumn{10}{c}{SwinV2} \\
    \midrule
    B   & \textbf{99.75} & \textbf{97.70} & \textbf{1.94} & 58.99 & \underline{57.36} & 45.05 & 59.25 & \underline{58.32} & 43.80 \\
    T-H & 99.29 & 96.59 & 3.40 & 57.60 & 54.96 & 44.79 & \underline{60.96} & 56.67 & \underline{42.21} \\
    T-HS& 99.49 & \underline{97.56} & 2.28 & \textbf{61.51} & \textbf{58.40} & \textbf{42.54} & \textbf{62.58} & \textbf{60.04} & \textbf{40.63} \\
    NT  & \underline{99.63} & 97.45 & \underline{1.96} & \underline{59.67} & 57.16 & \underline{43.64} & 60.18 & \underline{58.32} & 42.50 \\
    SC  & 99.61 & 97.03 & 2.36 & 58.77 & 54.91 & 44.44 & 58.43 & 57.45 & 43.77 \\
    \midrule
    \multicolumn{10}{c}{Efficient ViT} \\
    \midrule
    B & \textbf{99.71} & \underline{98.96} & \underline{1.97} & 55.61 & 55.19 & 46.01 & 57.98 & 57.55 & \underline{45.03} \\
    T-H & \underline{99.22} & \textbf{99.09} & \textbf{1.95} & \textbf{61.95} & \textbf{58.87}  & \textbf{43.92} & \textbf{61.70} & \textbf{62.31} & \textbf{40.32} \\
    T-HS & 97.15 & 82.05 & 11.71 & 51.91 & 50.02 & 50.11 & 52.08 & 51.99 & 49.71 \\
    NT & 98.92 & 98.23 & 4.61 & \underline{55.89} & \underline{56.06} & 48.24 & \underline{58.36} & \underline{59.05} & 45.92 \\
    SC & 98.73 & 98.11 & 5.70 & 54.52 & 54.94 & \underline{47.34} & 56.08 & 56.77 & 46.03 \\
    \bottomrule
  \end{tabularx}
\end{table*}

Furthermore, we aggregate the results of RQ2 to compare the performance of CNN-based models (EfficientNet, ConvNeXt) and Vision Transformer (ViT)-based models (Pyramid Vision Transformer, Swin) in terms of classification accuracy. We also include the aggregated accuracy of Efficient ViT, representing a hybrid architecture that combines CNN and ViT components. The results are summarized in Table~\ref{tab:cnn_vit_comparison}. Across most scenarios and manipulation types, CNNs consistently outperform ViTs. A notable exception arises in the within-dataset evaluation for the FaceShifter and FaceSwap manipulations, where ViTs slightly outperform CNNs. Efficient ViT also performs competitively in some within-dataset cases, occasionally surpassing  CNNs. However, in the cross-dataset setting, it slightly lags behind CNNs, though the difference is marginal.

Regarding RQ3, which examines whether contrastive methods enhance the generalization ability of DeepFake detectors, the results are summarized in Table \ref{tab:rq3}. A first observation is that, in the within-dataset setting, the inclusion of contrastive loss does not lead to performance improvements, likely due to the already high baseline performance. In the cross-dataset scenario, minimal gains are observed for the smaller networks like EfficientNetV2 and PVT-V2 across all metrics, with a slight improvement for PVT when tested on CelebDF. In contrast, the Swin model outperforms the baseline in terms of generalization to CelebDF and DFDC, mainly with the use of triplet loss with hard positive and semi-hard negative mining. Conversely, while ConvNext performs below the baseline on CelebDF, it shows significant improvement on DFDC with the addition of Supervised Contrastive loss. In the case of Efficient ViT, triplet loss with hard mining yields the highest performance gains on both CelebDF and DFDC datasets, while the NT-Xent loss consistently ranks among the second-best performing methods. These results indicate that a well-designed contrastive formulation could potentially lead to superior performance. In addition, contrastive learning could help address distribution shifts by forming sample pairs across different datasets, forcing models to learn features invariant to variations in capturing conditions. 

\section{Conclusions} 
\label{conclusions}
In this work, we aimed to investigate the generalization capabilities of DeepFake attribution models and explore whether the use of contrastive loss can lead to performance improvements. Overall our findings support the two following conclusions. First, DeepFake attribution models with vanilla approaches generalize less compared to binary counterparts. Incorporating contrastive methods is helpful, especially in larger models such as ViT and CNN + ViT combinations. Second, the ability of attribution models to maintain their accuracy across datasets is heavily influenced by data quality. Training on high-quality DeepFakes drastically improves the overall performance. Yet, our study indicates that deploying DeepFake detection in the wild is likely to lead to highly unreliable detection, especially in cases where the tested cases deviate significantly in terms of characteristics from the training sets. 
Our plans for future work include the use of spatiotemporal models, such as Video Foundation models, in order to investigate how the temporal dimension affects performance, how well these models generalize across datasets, and whether large-scale foundation models can lead to improvements in generalization.

\begin{acks}
This work is partially funded by the Horizon Europe project \href{https://www.veraai.eu/}{vera.ai} under grant agreement no. 101070093 and \href{https://aicode-project.eu/}{AI-CODE} under grant agreement no. 101135437.
\end{acks}

\balance

\clearpage

\bibliographystyle{ACM-Reference-Format}
\bibliography{refs}

\balance

\end{document}